\documentclass{article}

\usepackage{arxiv}
\usepackage{amsmath}
\renewcommand{\today}{April 2024}
\usepackage[utf8]{inputenc} 
\usepackage[T1]{fontenc}    
\usepackage{hyperref}       
\usepackage{url}            
\usepackage{booktabs}       
\usepackage{amsfonts}       
\usepackage{nicefrac}       
\usepackage{microtype}      
\usepackage{lipsum}
\usepackage{graphicx}
\usepackage{float}
\graphicspath{ {./images/} }

\title{When Metrics Disagree: Automatic Similarity vs. LLM-as-a-Judge for Clinical Dialogue Evaluation}

\makeatletter
\def\@date{April 2024}
\makeatother

\author{
  \textbf{Bian Sun} \\
  Heinz College \\
  Carnegie Mellon University \\
  \texttt{bians@andrew.cmu.edu}
  \and
  \textbf{Zhenjian Wang} \\
  Heinz College \\
  Carnegie Mellon University \\
  \texttt{zhenjiaw@andrew.cmu.edu}
  \vspace{10pt} %
  \and
  \textbf{Orvill de la Torre} \\
  School of Computer Science \\
  Carnegie Mellon University \\
  \texttt{odelator@andrew.cmu.edu}
  \and
  \textbf{Zirui Wang} \\
  Heinz College \\
  Carnegie Mellon University \\
  \texttt{ziruiw2@andrew.cmu.edu}
}
\date{\textbf{April 2024}} 

\begin{document}

\maketitle

\begin{abstract}
\noindent
\leftskip=1cm 
\rightskip=1cm 
As Large Language Models (LLMs) are increasingly integrated into healthcare to address complex medical inquiries, ensuring their reliability remains a critical challenge. Recent studies, including those from Long Island University, have highlighted that generic LLMs often struggle in clinical contexts, occasionally producing misleading or harmful guidance. To mitigate these risks, this research focuses on the domain-specific adaptation of \textbf{Llama-2-7B}, a transformer-based, decoder-only architecture, using the \textbf{Low-Rank Adaptation (LoRA)} technique. By injecting trainable low-rank matrices into the Transformer layers, we efficiently adapted the model using authentic patient-physician conversational transcripts while preserving the foundational knowledge of the base model. Our objective was to enhance the model's precision and contextual relevance in responding to medical queries by capturing the specialized nuances of clinical discourse. 

Due to the resource-intensive nature of large-scale human expert validation, the performance of the fine-tuned model was evaluated through a dual-track framework: \textbf{Track A} utilized traditional lexical similarity metrics (e.g., BLEU, ROUGE), while \textbf{Track B} employed an "LLM-as-a-Judge" paradigm using GPT-4 for semantic assessment. Our results demonstrate that while the LoRA-enhanced model achieved statistically significant improvements across all quantitative lexical dimensions, a profound disagreement surfaced in the GPT-4 evaluation, which marginally favored the baseline model's conversational flow. This metric divergence underscores a pivotal finding: traditional automated scores may not fully reflect clinical utility. Consequently, we propose that while automated metrics and LLM judges serve as valuable developmental proxies, rigorous validation by human medical experts remains an indispensable requirement for the safe deployment of LLMs in healthcare settings.
\end{abstract}

\section{Introduction}
In the rapidly evolving landscape of Natural Language Processing (NLP), Large Language Models (LLMs) have emerged as a cornerstone technology for understanding and generating human-like text. At the forefront of this revolution is Llama, a foundational open-source model released by Meta AI. Characterized by its multi-billion parameter scale, Llama \cite{touvron2023llama2openfoundation} leverages a refined decoder-only Transformer architecture that has demonstrated extraordinary efficiency in a variety of downstream tasks. Unlike closed-source alternatives, the open-source nature of Llama enables researchers to delve into its source code, facilitating the implementation of domain-specific fine-tuning—a process that adjusts model parameters by introducing specialized training data to align the model with high-stakes fields, such as clinical medicine.

We selected the Llama-2-7B variant as our baseline due to its optimal balance between computational feasibility and generative performance. Llama's architecture incorporates several sophisticated innovations: data scaling is managed via \cite{zhang2019rootmeansquarelayer} (Root Mean Square Layer Normalization), while the traditional ReLU activation function is substituted with  \cite{shazeer2020gluvariantsimprovetransformer} to enhance non-linear representation. Furthermore, absolute positional embeddings are replaced by Rotary Positional Embeddings (RoPE) to improve sequence flexibility. For optimization, Llama utilizes the AdamW optimizer ($\beta_1 = 0.9, \beta_2 = 0.95$) paired with a cosine learning rate scheduler, weight decay of 0.1, and gradient clipping to ensure training stability.

The adoption of Llama offers distinct strategic advantages. Its self-supervised training on massive unlabeled datasets eliminates the need for labor-intensive manual annotation, allowing the model to internalize vast linguistic patterns. Compared to proprietary models like ChatGPT, Llama features a more streamlined architecture that requires fewer computational resources, making it highly accessible for both commercial and academic research. Its versatility makes it ideal for applications such as text categorization, automated summarization, and entity recognition. However, certain limitations persist; the 7B parameter scale may constrain complex creative writing or highly nuanced narrative generation. Additionally, while efficient, these models still demand significant storage and can exhibit fragile generalization when applied to domains far removed from their pre-training distribution.

The potential of LLMs is particularly transformative in the medical sector, where the accuracy and reliability of information are paramount. Recent advancements have shown that models like Llama, when fine-tuned on clinical datasets, can improve diagnostic accuracy and patient interactions. Our project contributes to this effort by fine-tuning Llama specifically for patient-doctor dialogues in real-world scenarios. Leveraging a robust dataset of 5,000 conversational exchanges and a comprehensive disease database, we utilized Low-Rank Adaptation (LoRA)\cite{hu2021loralowrankadaptationlarge} to enhance the model's clinical intuition.

Crucially, this research explores the Evaluation Paradox inherent in specialized AI. While our experimental results indicate that the fine-tuned model surpasses the "vanilla" baseline across all traditional lexical metrics (Track A: BLEU, ROUGE, and METEOR) \cite{banerjee-lavie-2005-meteor,lin-2004-rouge, Papineni2002BleuAM}, we observed a profound disagreement when employing GPT-4 as a Judge (Track B)\cite{zheng2023judgingllmasajudgemtbenchchatbot}. This divergence—where automatic similarity metrics and LLM-based semantic evaluations provide conflicting conclusions—exposes a critical gap in current AI validation. Through this study, we highlight that while automated metrics are essential developmental proxies, they cannot replace the necessity for human expert evaluation in ensuring patient safety and diagnostic reliability in healthcare applications.

\section{Related Work}

\label{gen_inst}

\subsection{Evolution of Neural Language Modeling}
The progression toward Large Language Models (LLMs) originates from fundamental research in deep learning applied to sequence modeling. Early architectures utilized Recurrent Neural Networks (RNNs) and Long Short-Term Memory (LSTM) networks to address the challenges of vanishing gradients and long-range dependencies in natural language \cite{schmidt2019recurrentneuralnetworksrnns,staudemeyer2019understandinglstmtutorial}. However, the introduction of the Transformer architecture \cite{vaswani2023attentionneed} catalyzed a paradigm shift by employing self-attention mechanisms. This innovation allowed for unprecedented scalability and contextual understanding, moving beyond the sequential constraints of previous models and enabling the development of massive pre-trained models that define the current state-of-the-art.

\subsection{The Llama Framework and Open-Source Adaptation}
Meta AI's release of Llama represented a significant milestone by providing a high-performance, open-source alternative to proprietary models like ChatGPT. Ranging from 7B to 65B parameters, Llama was designed for versatility and efficiency \cite{touvron2023llama2openfoundation}. Research has consistently demonstrated that Llama achieves comparable fluency and coherence to larger models while maintaining lower computational complexity. Its open-source nature has proven particularly advantageous for the research community, as it allows for direct architectural intervention and parameter-efficient fine-tuning (PEFT) in specialized domains where the source code of closed-source counterparts remains inaccessible.

\subsection{LLMs in Clinical Medicine and Healthcare}
The application of LLMs in the medical domain has seen rapid growth, with several key studies benchmarking the potential of model adaptation:

\textbf{Medical Dialogue Systems:} Li et al. \cite{li2023chatdoctormedicalchatmodel} developed ChatDoctor (MediDialogue-LLM), a conversational agent fine-tuned for diagnostic assistance and patient interaction. Their findings underscored the model's ability to handle complex clinical queries but also highlighted a persistent need for professional oversight due to occasional generative inaccuracies.

\textbf{Medical Licensing and Knowledge}: Cerebras Systems utilized high-performance computing to adapt Llama-2-70B for medical examinations. The resulting model, Med42, achieved a significant 72\% score on the USMLE, demonstrating that large-scale clinical datasets and computational scaling can push open-source models beyond expert-level benchmarks \cite{christophe2024med42v2suiteclinicalllms}.

\textbf{Administrative and Diagnostic Optimization: }Wang et al. \cite{wang_gao_dantona_hull_sun_2024} explored the utility of Llama-7B for Diagnosis-Related Group (DRG) assignment within the U.S. inpatient payment system. Despite limitations in data and computational resources, their work indicated that LLMs could significantly improve the efficiency of automated clinical coding and administrative workflows.

\section{Method}
\subsection{Baseline Selection}
\label{headings}

For our baseline architecture, we selected Llama-2-7B \cite{touvron2023llama2openfoundation}, a highly capable large language model optimized for robust dialogue and text generation. To illustrate the foundational conversational capabilities of our baseline model prior to clinical evaluation, we documented its responses to various prompts. Figure 1 showcases a sample interaction generated during this initial testing phase.

\begin{figure}[h]
    \centering
    \includegraphics[width=1\textwidth]{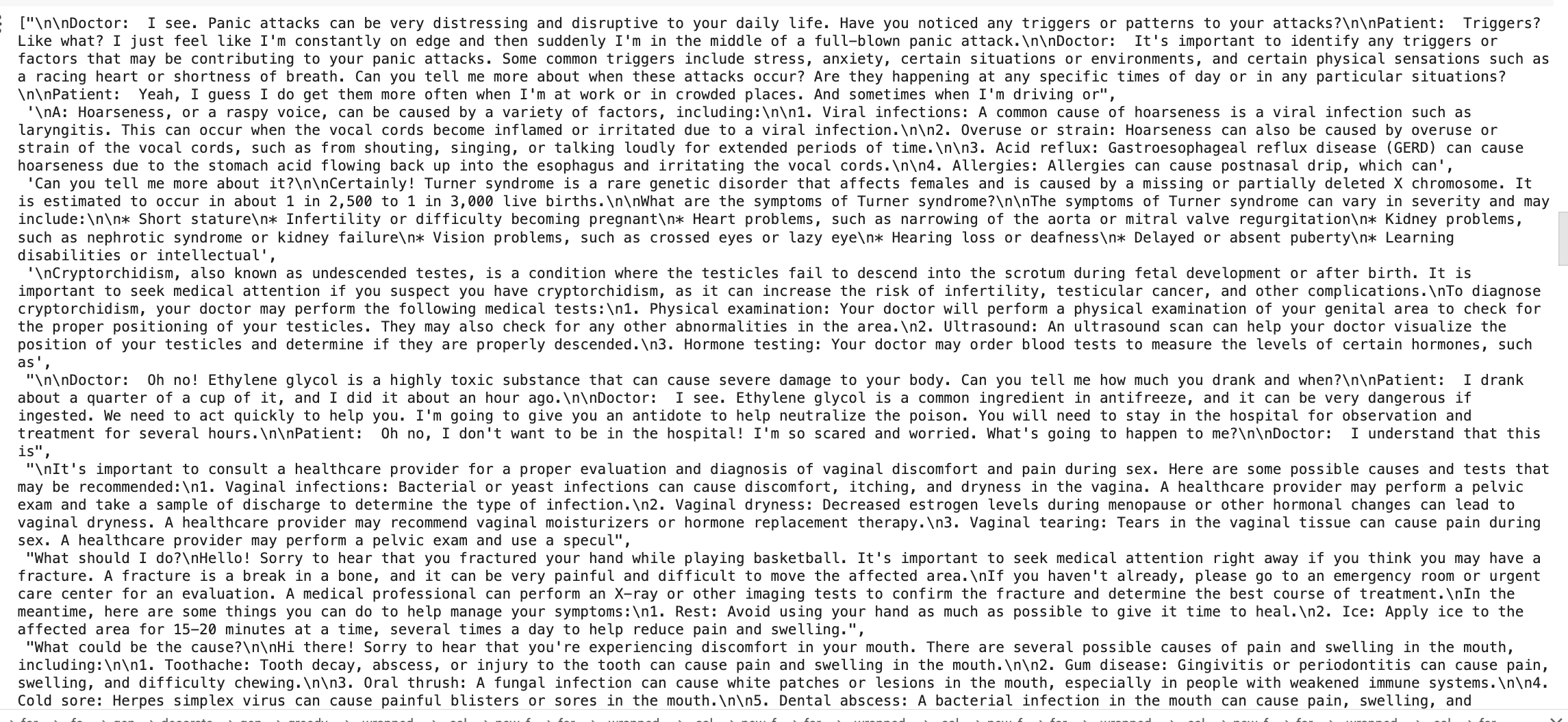}
    \caption{Our Execution Output of the Baseline Model}
\end{figure}

The Llama architecture operates on a decoder-only transformer framework. To stabilize training, it utilizes Root Mean Square Normalization (RMSNorm) \cite{zhang2019rootmeansquarelayer}at the input of each transformer sub-layer, replacing standard layer normalization. Additionally, Llama replaces the traditional ReLU non-linearity with the SwiGLU \cite{shazeer2020gluvariantsimprovetransformer} activation function, which has empirically demonstrated superior performance. SwiGLU is a fusion of the Swish and Gated Linear Unit (GLU) activation functions, designed to leverage their combined strengths. Mathematically, it is defined as:

\begin{equation}
    \text{SwiGLU}(x) = x \cdot \sigma(\beta \cdot x) + (1 - \sigma(\beta \cdot x)) \cdot (Wx + b)
\end{equation}

where $x$ represents the input, and $W$, $b$, and $\beta$ are trainable parameters. By employing the Swish function to gate the linear function of the GLU, SwiGLU effectively captures complex, non-linear patterns in the data while maintaining computational efficiency. SwiGLU offers several distinct advantages over alternative activation functions, primarily due to its smoothness, non-monotonicity, and gating mechanism. First, it provides smoother gradients than ReLU, facilitating better optimization and faster convergence. Second, its non-monotonic nature enables the network to learn highly intricate relationships between inputs and outputs. Finally, the gating mechanism allows for the selective activation of neurons based on the input context, which helps reduce overfitting and enhances generalization. Consistent with our observations in conversational modeling, empirical studies demonstrate that SwiGLU outperforms both standard Swish and GLU across a variety of domains, including advanced language modeling and machine translation.

\begin{figure}[h]
    \centering
    \includegraphics[width=.75\textwidth]{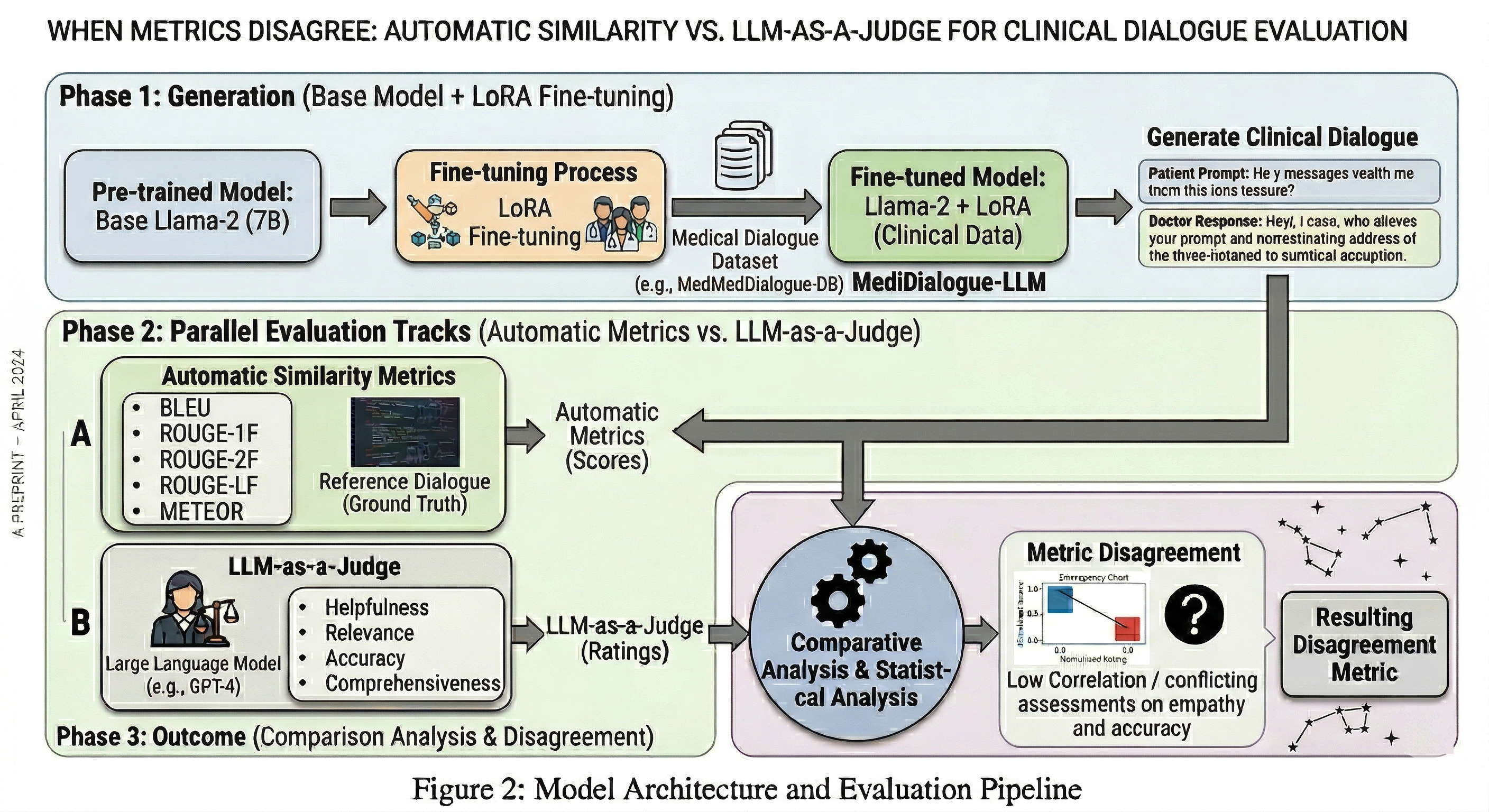}
    \caption{Model Architecture and Evaluation Pipeline}
\end{figure}

\subsection{Domain Adaptation for Clinical Dialogue via LoRA}
To generate the specialized medical responses required for our comparative evaluation pipeline, we adapted the baseline Llama-2 model into our clinical dialogue system, MediDialogue-LLM. To optimize this system efficiently, we employed Low-Rank Adaptation (LoRA) \cite{hu2021loralowrankadaptationlarge}, a highly effective Parameter-Efficient Fine-Tuning (PEFT) \cite{xu2023parameterefficientfinetuningmethodspretrained} methodology. LoRA achieves downstream model improvement with minimal computational overhead, proving functionally comparable to full fine-tuning while significantly lowering the economic and hardware barriers to LLM development.

LoRA operates on the premise that the weight matrices of pre-trained LLMs possess a low intrinsic dimension, meaning they often contain substantial redundancies. Rather than updating the entire parameter space during fine-tuning, LoRA streamlines the process by freezing the pre-trained model weights and injecting trainable rank-decomposition matrices into the Transformer architecture.

Mathematically, the LoRA-based fine-tuning process is defined as:

\begin{equation}
W = W_o + \Delta W = W_o + BA
\end{equation}

where $W_0 \in \mathbb{R}^{d \times k}$ denotes the frozen pre-trained parameter weights, $\Delta W$ represents the learned weight updates, and $W$ is the final weight matrix utilized during inference. The update $\Delta W$ is factorized into two matrices: $B \in \mathbb{R}^{d \times r}$ and $A \in \mathbb{R}^{r \times k}$. By restricting the rank such that $r \ll \min(d, k)$, this approach drastically reduces the number of trainable parameters.Fine-tuning with LoRA offers several critical advantages for our study. It inherently reduces memory and storage constraints, mitigates the risk of catastrophic forgetting of the model's foundational language capabilities, and introduces no additional inference latency. Consequently, it provides an efficient and practical mechanism for adapting pre-trained LLMs to the specialized clinical domain \cite{li2023chatdoctormedicalchatmodel}. This ensures that the generated medical dialogues are both contextually accurate and linguistically fluent, thereby providing a robust testing ground for our core investigation: comparing automatic similarity metrics against LLM-as-a-judge paradigms.

\subsection{Fine-Tuning Objective}

During the LoRA adaptation phase, we optimize the ChatDoctor (Llama-2) model using the standard causal language modeling objective. Consistent with the model's pre-training stage, the parameter updates are driven by minimizing the cross-entropy loss over the generated clinical responses. 

In the context of autoregressive language generation, the cross-entropy loss evaluates the divergence between the true token distribution and the model's predicted probability distribution over the vocabulary. For a given clinical dialogue sequence \( X = (x_1, x_2, ..., x_T) \), the model is trained to predict the next token \( x_t \) given the preceding context \( x_{<t} \). The loss function \( \mathcal{L} \) is mathematically defined as:

\begin{equation}
\mathcal{L} = - \frac{1}{T} \sum_{t=1}^{T} \log P(x_t | x_{<t}; \Theta)
\end{equation}

where:
\begin{itemize}
  \item \( T \) represents the total number of tokens in the generated target sequence.
  \item \( x_t \) is the true target token at time step \( t \).
  \item \( x_{<t} \) denotes the contextual sequence of all preceding tokens.
  \item \( P(x_t | x_{<t}; \Theta) \) is the conditional probability of generating the correct token \( x_t \), as estimated by the model with parameters \( \Theta \).
\end{itemize}

By iteratively minimizing this cross-entropy loss during the LoRA fine-tuning process, the model effectively maximizes the likelihood of generating contextually accurate and medically relevant tokens, thereby aligning its underlying language representations with the specialized clinical domain.

\section{Experimental Setup}
\subsection{Dataset Description}
\label{headings}

To establish a controlled baseline for our fine-tuning and evaluation pipeline, we primarily utilized the GenMedGPT-5k dataset. This curated corpus consists of 5,000 synthetic patient-physician conversational exchanges, generated via ChatGPT guided by a structured medical disease database to ensure clinical relevance.\footnote{The GenMedGPT-5k dataset is accessible via the ChatDoctor repository: \url{https://github.com/Kent0n-Li/ChatDoctor}} 

While our long-term objective includes scaling this evaluation framework to larger, real-world datasets—such as the HealthCareMagic-100k corpus, which contains 100,000 patient-doctor dialogues from an online consultation platform\footnote{Available at: \href{https://huggingface.co/datasets/wangrongsheng/HealthCareMagic-100k-en}{Hugging Face - HealthCareMagic-100k}}—the current scope of this study strictly focuses on the GenMedGPT-5k corpus. This smaller, highly structured dataset provides a noise-free environment essential for rigorously observing the divergence between automatic similarity metrics and LLM-based judgments.

Furthermore, to operationalize our LLM-as-a-Judge framework (Track B), we compiled the structured evaluation outputs generated by GPT-4. These GPT-4 assessments form the comparative basis against our traditional metric scores (Track A).

\subsection{Evaluation Metrics}
\subsubsection{Track A: Automatic Similarity Metrics}
\textbf{Bilingual Evaluation Understudy (BLEU):
}BLEU \cite{Papineni2002BleuAM}is a popular metric for evaluating the quality of text which has been machine-translated from one natural language to another. It measures the correspondence between a machine's output and that of a human. Commonly, BLEU scores are calculated based on n-gram precision, with a penalty for too-short translations. The BLEU score is computed as follows:
\begin{equation}
\text{BLEU} = BP \cdot \exp\left(\sum_{n=1}^N w_n \log p_n\right)
\end{equation}
where \(p_n\) is the precision of n-grams, \(w_n\) are weights summing to 1, \(N\) is the length of n-gram, and \(BP\) is the brevity penalty to penalize short machine-generated translations.

\textbf{Recall-Oriented Understudy for Gisting Evaluation - Unigram F1 Score (ROUGE-1F):}
ROUGE-1F \cite{lin-2004-rouge}measures the overlap of unigrams between the system-generated output and reference texts. It is used particularly in the evaluation of summary texts. The F1 score is the harmonic mean of precision and recall, calculated as follows:
\begin{equation}
\text{ROUGE-1F} = \frac{2 \times (\text{Precision} \times \text{Recall})}{\text{Precision} + \text{Recall}}
\end{equation}
where Precision is the ratio of the number of overlapping unigrams in generated and reference texts to the total number of unigrams in the generated text, and Recall is the ratio of overlapping unigrams to the total number of unigrams in the reference text.

\textbf{Recall-Oriented Understudy for Gisting Evaluation - Bigram F1 Score (ROUGE-2F):}
ROUGE-2F assesses the overlap of bigrams between the system output and a set of reference translations. It is useful for evaluating the coherence of generated texts. The equation for ROUGE-2F, similar to ROUGE-1F, is:
\begin{equation}
\text{ROUGE-2F} = \frac{2 \times (\text{Precision} \times \text{Recall})}{\text{Precision} + \text{Recall}}
\end{equation}
This score considers bigrams, thus providing insight into the fluency of the text. \(P\) (precision) and \(R\) (recall) are measures of matching unigrams between the machine output and human reference.

\textbf{Recall-Oriented Understudy for Gisting Evaluation - Longest Common Subsequence F-measure (ROUGE-LF)}:
ROUGE-LF is a metric specifically designed to evaluate the quality of text summarization by measuring the longest common subsequence between the system-generated output and the reference texts. Unlike other metrics that may consider only exact matches, ROUGE-LF appreciates the order of words, making it particularly sensitive to the fluency and flow of the generated text. The ROUGE-LF score is formulated as:
\begin{equation}
\text{ROUGE-LF} = \frac{(1 + \beta^2) \cdot P \cdot R}{R + \beta^2 \cdot P}
\end{equation}
where \(P\) (precision) is the proportion of the longest common subsequence words in the generated text that are also in the reference text, \(R\) (recall) is the proportion of the longest common subsequence words in the reference that are also in the generated text, and \(\beta\) is typically set to favor recall (e.g., \(\beta = 1.2\)) to stress the importance of capturing more of the reference content.

\textbf{Metric for Evaluation of Translation with Explicit ORdering (METEOR)}:
METEOR \cite{banerjee-lavie-2005-meteor}is another metric for evaluating translation output quality. Unlike BLEU, METEOR considers synonym matches and stems alongside exact word correspondences, and also incorporates structure into its evaluation. The METEOR score is calculated as follows:
\begin{equation}
\text{METEOR} = \frac{10 \cdot P \cdot R}{R + 9 \cdot P}
\end{equation}
where \(P\) (precision) and \(R\) (recall) are measures of matching unigrams between the machine output and human reference.

\subsubsection{Track B: LLM-as-a-Judge}

While human expert evaluation by clinical professionals represents the gold standard for assessing medical dialogue, the resource-intensive nature of manual scoring inherently limits the scalability of large-scale evaluations. To overcome this limitation and enable a comprehensive analysis of our generated dialogues, we adopted the "LLM-as-a-Judge" \cite{zheng2023judgingllmasajudgemtbenchchatbot}. Recent literature demonstrates that strong large language models, when guided by rigorously defined rubrics, exhibit high agreement with human domain experts. Following this methodology, we employed GPT-4 as an automated evaluator to assess the quality of the model responses across four critical clinical dimensions:

\textbf{Helpfulness:} Measures the extent to which the generated text practically aids the patient in achieving their specific health-related goals or resolving their medical inquiry.

\textbf{Relevance:} Assesses the strict alignment of the generated response with the specific symptoms, queries, or clinical concerns posed by the user, penalizing off-topic or generic generation.

\textbf{Accuracy:} Evaluates the factual and medical correctness of the information provided, ensuring that the model's advice is scientifically valid, safe, and aligned with current medical consensus.

\textbf{Comprehensiveness:} Refers to the depth and breadth of the clinical information provided. A highly comprehensive output covers all essential aspects and nuances of the user's query without overwhelming the patient with tangential information or overly dense medical jargon.

\subsection{Implemented Details}
To generate the domain-specific clinical responses required for our comparative metric analysis, we implemented a fine-tuning pipeline optimized for an NVIDIA A100 GPU environment. The foundational architecture selected was the instruction-tuned \texttt{llama-2-7b-chat-hf} variant. 

The domain adaptation was executed using the LoRA (Low-Rank Adaptation) methodology to ensure parameter efficiency. The model was trained on the previously detailed 5,000-sample clinical dataset, with 120 randomly sampled dialogue pairs strategically withheld as the validation set to monitor generalization and prevent overfitting. 

To stabilize the training process within our computational constraints, we established the following hyperparameter configuration:
\begin{itemize}
    \item \textbf{Learning Rate:} $3 \times 10^{-5}$. This conservative value was selected to facilitate clinical domain adaptation while mitigating the risk of catastrophic forgetting of the base model's conversational capabilities.
    \item \textbf{Epochs and Batching:} The optimization was conducted over a single training epoch (\( \text{num\_epochs} = 1 \)), utilizing a strict batch size and micro-batch size of 1.
    \item \textbf{Sequence Length:} The maximum input sequence length (\texttt{cutoff\_len}) was capped at 256 tokens to optimize memory allocation during the gradient update phase.
\end{itemize}

\section{Results}
\subsection{Automatic Evaluation Results} 
To quantify the impact of clinical domain adaptation, we first evaluated the model outputs using our suite of automatic similarity metrics: BLEU, ROUGE-1F, ROUGE-2F, ROUGE-LF, and METEOR. For this evaluation, we utilized a filtered evaluation set of $N=84$ complete, well-formed patient-doctor dialogue exchanges, explicitly excluding truncated outputs or responses impacted by weight-merging latency to ensure a fair comparison.

Table 1 summarizes the performance comparison between the baseline Llama-2 model and our fine-tuned clinical model. The results demonstrate that the fine-tuned model achieves strict superiority across all evaluated metrics when compared against the ground-truth reference responses.

Notably, the BLEU and ROUGE-2F indices exhibited the most pronounced sensitivity to the fine-tuning process. The fine-tuned model achieved a massive 548.99\% relative improvement in BLEU score and an 84.35\% relative improvement in ROUGE-2F. These specific gains indicate that the domain adaptation successfully taught the model not just to use the correct clinical vocabulary (captured by unigram metrics), but to sequence those terms into coherent, exact-match phrases that closely mirror professional medical discourse. 

To validate the robustness of these improvements, we conducted a non-parametric Mann-Whitney U test across all score distributions. As detailed in Table 1, the $p$-values for all five automatic metrics fall well below the standard threshold of statistical significance ($p < 0.05$). This confirms that the lexical and structural enhancements observed in the fine-tuned model's replies are statistically distinct from the baseline, establishing a strong, quantitatively validated foundation before introducing our semantic LLM-as-a-Judge evaluation.

\begin{table}[htbp]
\centering
\caption{Performance Comparison of Models across Automatic Similarity Metrics}
\label{tab:model_comparison}
\begin{tabular}{@{}lccccc@{}}
\toprule
\textbf{Model} & \textbf{BLEU} & \textbf{ROUGE-1F} & \textbf{ROUGE-2F} & \textbf{ROUGE-LF} & \textbf{METEOR} \\ 
\midrule
Llama-2 (Baseline) & 0.0033 & 0.1819 & 0.0298 & 0.1696 & 0.1004 \\
Fine-tuned Model   & 0.0216 & 0.2211 & 0.0549 & 0.2061 & 0.1227 \\ 
\midrule
\textbf{Difference} & +0.0183 & +0.0392 & +0.0251 & +0.0365 & +0.0223 \\
\textbf{Improvement (\%)} & 548.99\% & 21.54\% & 84.35\% & 21.56\% & 22.19\% \\ 
\textbf{$p$-value} & 0.0040 & 0.0070 & 0.0162 & 0.0100 & 0.0450 \\
\bottomrule
\end{tabular}
\end{table}

\subsection{Track B Results: LLM-as-a-Judge (GPT-4) Evaluation}

To conduct our semantic evaluation framework (Track B), we randomly sampled a subset of $N=20$ clinical queries and extracted the corresponding responses generated by both the baseline Llama-2 and the fine-tuned MediDialogue-LLM. We then employed GPT-4 to blindly score these responses on a scale of 1 to 5 across the four previously defined dimensions: Accuracy, Comprehensiveness (Detail), Helpfulness, and Relevance.

The aggregate results of this evaluation are presented in Table \ref{tab:gpt4_evaluation}, reporting the mean scores for each model alongside the corresponding $p$-values. 

\begin{table}[htbp]
\centering
\caption{GPT-4 Evaluation Results (Mean Scores out of 5, $N=20$)}
\label{tab:gpt4_evaluation}
\begin{tabular}{lccc}
\toprule
\textbf{Evaluation Dimension} & \textbf{Llama-2 (Baseline)} & \textbf{Fine-tuned Model} & \textbf{$p$-value} \\ 
\midrule
Accuracy & \textbf{2.80} & 2.65 & 0.6868 \\
Comprehensiveness & \textbf{3.55} & 3.30 & 0.4026 \\ 
Helpfulness & \textbf{3.25} & 3.15 & 0.7605 \\
Relevance & \textbf{2.75} & 2.60 & 0.6889 \\
\bottomrule
\end{tabular}
\end{table}

Strikingly, the results from the LLM-as-a-Judge evaluation completely diverge from the conclusions drawn by the automatic similarity metrics in Track A. While BLEU and ROUGE metrics heavily favored the fine-tuned model, GPT-4 assigned marginally higher mean scores to the baseline Llama-2 across all four semantic dimensions. However, subsequent statistical testing revealed that all $p$-values significantly exceed the 0.05 threshold, indicating that the performance difference between the two models from GPT-4's perspective is not statistically significant.

\subsection{The Disagreement: Why Human Evaluation Remains Indispensable}

The conflicting outcomes between Track A and Track B expose a critical vulnerability in current automated evaluation paradigms for clinical dialogue. 

Quantitative metrics like BLEU and ROUGE (Track A) are highly sensitive to surface-level lexical overlap. The massive score improvements observed in the fine-tuned model suggest it successfully memorized and replicated specific medical terminology and phrasing structures from the training data. Conversely, the GPT-4 evaluator (Track B) prioritizes narrative flow, conversational coherence, and general semantic alignment. The baseline model's slight edge in these scores implies that while the fine-tuned model learned domain-specific vocabulary, the base model may still retain a more generalized, fluent conversational ability that the LLM judge inherently prefers. 

More importantly, neither automated approach possesses genuine clinical expertise. Automatic metrics mechanically reward exact word matches without understanding medical context, while the GPT-4 judge—despite its advanced reasoning capabilities—evaluates medical accuracy and helpfulness based on probabilistic language generation rather than rigorous medical training. 

Therefore, we conclude that while automatic similarity metrics and LLM judges are valuable for establishing baseline functional differences during model development, their profound disagreement underscores their collective inadequacy for ultimate clinical validation. To ensure patient safety, mitigate medical hallucination, and truly assess the diagnostic reliability of medical LLMs, gold-standard evaluation by human medical professionals remains indispensable. \\

\section{Conclusion}
This study investigated the efficacy of domain-specific fine-tuning for clinical dialogue generation and, crucially, the reliability of the paradigms used to evaluate it. By applying Low-Rank Adaptation (LoRA) to the foundational Llama-2-7B model, we developed a specialized model optimized for medical inquiries. Our initial quantitative analysis demonstrated that the fine-tuned model vastly outperformed the baseline across all traditional lexical similarity metrics, achieving statistically significant improvements in BLEU, ROUGE, and METEOR scores.

However, our semantic evaluation using the LLM-as-a-Judge paradigm (GPT-4) revealed a stark divergence: the automated judge favored the conversational outputs of the baseline model across all qualitative dimensions, including accuracy, comprehensiveness, and helpfulness.

This profound disagreement between evaluation tracks forms the core insight of our research. It exposes the intrinsic limitations of relying solely on surface-level n-gram metrics, which blindly reward the replication of clinical vocabulary but fail to capture semantic coherence, conversational fluency, or genuine medical helpfulness. Conversely, it highlights that even advanced LLM evaluators may prioritize generalized conversational flow over strict, domain-adapted clinical precision. Ultimately, our findings underscore a critical reality in medical AI development: while automated metrics are valuable developmental proxies, they possess inherent biases. To ensure diagnostic reliability and patient safety, the deployment of medical LLMs continues to necessitate rigorous, gold-standard validation by human clinical experts.
 Here use ChatGPT4 evaluation, the results of the ChatGPT 4 are quite different from the evaluation metrics. It may need more data and computer resources to  give a deeper research. And here we suggest that, the human evaluation is very important, because this two evaluations did not have the same outcome. So we need more time to find what are the reasons which lead to the difference.

These findings are encouraging as they doubt the effectiveness of fine-tuning in transforming LLMs into specialized tools across various industries. Our methodology offers a replicable framework for other research teams interested in adapting models for enhanced performance in other specialized fields.

\bibliographystyle{unsrt} 
\bibliography{References} 

@misc{touvron2023llama2openfoundation,
      title={Llama 2: Open Foundation and Fine-Tuned Chat Models}, 
      author={Hugo Touvron and Louis Martin and Kevin Stone and Peter Albert and Amjad Almahairi and Yasmine Babaei and Nikolay Bashlykov and Soumya Batra and Prajjwal Bhargava and Shruti Bhosale and Dan Bikel and Lukas Blecher and Cristian Canton Ferrer and Moya Chen and Guillem Cucurull and David Esiobu and Jude Fernandes and Jeremy Fu and Wenyin Fu and Brian Fuller and Cynthia Gao and Vedanuj Goswami and Naman Goyal and Anthony Hartshorn and Saghar Hosseini and Rui Hou and Hakan Inan and Marcin Kardas and Viktor Kerkez and Madian Khabsa and Isabel Kloumann and Artem Korenev and Punit Singh Koura and Marie-Anne Lachaux and Thibaut Lavril and Jenya Lee and Diana Liskovich and Yinghai Lu and Yuning Mao and Xavier Martinet and Todor Mihaylov and Pushkar Mishra and Igor Molybog and Yixin Nie and Andrew Poulton and Jeremy Reizenstein and Rashi Rungta and Kalyan Saladi and Alan Schelten and Ruan Silva and Eric Michael Smith and Ranjan Subramanian and Xiaoqing Ellen Tan and Binh Tang and Ross Taylor and Adina Williams and Jian Xiang Kuan and Puxin Xu and Zheng Yan and Iliyan Zarov and Yuchen Zhang and Angela Fan and Melanie Kambadur and Sharan Narang and Aurelien Rodriguez and Robert Stojnic and Sergey Edunov and Thomas Scialom},
      year={2023},
      eprint={2307.09288},
      archivePrefix={arXiv},
      primaryClass={cs.CL},
      url={https://arxiv.org/abs/2307.09288}, 
}

@misc{zhang2019rootmeansquarelayer,
      title={Root Mean Square Layer Normalization}, 
      author={Biao Zhang and Rico Sennrich},
      year={2019},
      eprint={1910.07467},
      archivePrefix={arXiv},
      primaryClass={cs.LG},
      url={https://arxiv.org/abs/1910.07467}, 
}

@misc{shazeer2020gluvariantsimprovetransformer,
      title={GLU Variants Improve Transformer}, 
      author={Noam Shazeer},
      year={2020},
      eprint={2002.05202},
      archivePrefix={arXiv},
      primaryClass={cs.LG},
      url={https://arxiv.org/abs/2002.05202}, 
}

@misc{hu2021loralowrankadaptationlarge,
      title={LoRA: Low-Rank Adaptation of Large Language Models}, 
      author={Edward J. Hu and Yelong Shen and Phillip Wallis and Zeyuan Allen-Zhu and Yuanzhi Li and Shean Wang and Lu Wang and Weizhu Chen},
      year={2021},
      eprint={2106.09685},
      archivePrefix={arXiv},
      primaryClass={cs.CL},
      url={https://arxiv.org/abs/2106.09685}, 
}

@misc{xu2023parameterefficientfinetuningmethodspretrained,
      title={Parameter-Efficient Fine-Tuning Methods for Pretrained Language Models: A Critical Review and Assessment}, 
      author={Lingling Xu and Haoran Xie and Si-Zhao Joe Qin and Xiaohui Tao and Fu Lee Wang},
      year={2023},
      eprint={2312.12148},
      archivePrefix={arXiv},
      primaryClass={cs.CL},
      url={https://arxiv.org/abs/2312.12148}, 
}

@misc{li2023chatdoctormedicalchatmodel,
      title={ChatDoctor: A Medical Chat Model Fine-Tuned on a Large Language Model Meta-AI (LLaMA) Using Medical Domain Knowledge}, 
      author={Yunxiang Li and Zihan Li and Kai Zhang and Ruilong Dan and Steve Jiang and You Zhang},
      year={2023},
      eprint={2303.14070},
      archivePrefix={arXiv},
      primaryClass={cs.CL},
      url={https://arxiv.org/abs/2303.14070}, 
}

@inproceedings{Papineni2002BleuAM,
  title={Bleu: a Method for Automatic Evaluation of Machine Translation},
  author={Kishore Papineni and Salim Roukos and Todd Ward and Wei-Jing Zhu},
  booktitle={Annual Meeting of the Association for Computational Linguistics},
  year={2002},
  url={https://api.semanticscholar.org/CorpusID:11080756}
}

@inproceedings{lin-2004-rouge,
    title = "{ROUGE}: A Package for Automatic Evaluation of Summaries",
    author = "Lin, Chin-Yew",
    booktitle = "Text Summarization Branches Out",
    month = jul,
    year = "2004",
    address = "Barcelona, Spain",
    publisher = "Association for Computational Linguistics",
    url = "https://aclanthology.org/W04-1013/",
    pages = "74--81"
}

@inproceedings{banerjee-lavie-2005-meteor,
    title = "{METEOR}: An Automatic Metric for {MT} Evaluation with Improved Correlation with Human Judgments",
    author = "Banerjee, Satanjeev  and
      Lavie, Alon",
    editor = "Goldstein, Jade  and
      Lavie, Alon  and
      Lin, Chin-Yew  and
      Voss, Clare",
    booktitle = "Proceedings of the {ACL} Workshop on Intrinsic and Extrinsic Evaluation Measures for Machine Translation and/or Summarization",
    month = jun,
    year = "2005",
    address = "Ann Arbor, Michigan",
    publisher = "Association for Computational Linguistics",
    url = "https://aclanthology.org/W05-0909/",
    pages = "65--72"
}

@misc{zheng2023judgingllmasajudgemtbenchchatbot,
      title={Judging LLM-as-a-Judge with MT-Bench and Chatbot Arena}, 
      author={Lianmin Zheng and Wei-Lin Chiang and Ying Sheng and Siyuan Zhuang and Zhanghao Wu and Yonghao Zhuang and Zi Lin and Zhuohan Li and Dacheng Li and Eric P. Xing and Hao Zhang and Joseph E. Gonzalez and Ion Stoica},
      year={2023},
      eprint={2306.05685},
      archivePrefix={arXiv},
      primaryClass={cs.CL},
      url={https://arxiv.org/abs/2306.05685}, 
}

@misc{christophe2024med42v2suiteclinicalllms,
      title={Med42-v2: A Suite of Clinical LLMs}, 
      author={Clément Christophe and Praveen K Kanithi and Tathagata Raha and Shadab Khan and Marco AF Pimentel},
      year={2024},
      eprint={2408.06142},
      archivePrefix={arXiv},
      primaryClass={cs.CL},
      url={https://arxiv.org/abs/2408.06142}, 
}

@article{wang_gao_dantona_hull_sun_2024, title={DRG-LLaMA : tuning LLaMA model to predict diagnosis-related group for hospitalized patients}, volume={7}, ISSN={2398-6352}, DOI={10.1038/s41746-023-00989-3}, number={1}, journal={npj Digital Medicine}, publisher={npj Digital Medicine}, author={Wang, Hanyin and Gao, Chufan and Dantona, Christopher and Hull, Bryan and Sun, Jimeng}, year={2024} }

@misc{vaswani2023attentionneed,
      title={Attention Is All You Need}, 
      author={Ashish Vaswani and Noam Shazeer and Niki Parmar and Jakob Uszkoreit and Llion Jones and Aidan N. Gomez and Lukasz Kaiser and Illia Polosukhin},
      year={2023},
      eprint={1706.03762},
      archivePrefix={arXiv},
      primaryClass={cs.CL},
      url={https://arxiv.org/abs/1706.03762}, 
}

@misc{schmidt2019recurrentneuralnetworksrnns,
      title={Recurrent Neural Networks (RNNs): A gentle Introduction and Overview}, 
      author={Robin M. Schmidt},
      year={2019},
      eprint={1912.05911},
      archivePrefix={arXiv},
      primaryClass={cs.LG},
      url={https://arxiv.org/abs/1912.05911}, 
}

@misc{staudemeyer2019understandinglstmtutorial,
      title={Understanding LSTM -- a tutorial into Long Short-Term Memory Recurrent Neural Networks}, 
      author={Ralf C. Staudemeyer and Eric Rothstein Morris},
      year={2019},
      eprint={1909.09586},
      archivePrefix={arXiv},
      primaryClass={cs.NE},
      url={https://arxiv.org/abs/1909.09586}, 
}

\section*{Appendix}
\section{Discussion}
\label{headings}
\subsection{Relevance of Results}
Our project's primary aim was to enhance the Llama model for medical applications, particularly in handling patient inquiries with high accuracy and reliability. The fine-tuning process applied to the Llama model significantly improved its performance metrics such as BLEU, ROUGE, and METEOR scores. These enhancements are crucial as they directly correlate with the model's ability to understand and generate medically relevant responses.
\subsection{Sensitivity Analysis}
Our paper compares quantitative assessment methods with evaluations performed by ChatGPT-4. We observe the effectiveness and reliability of the fine-tuned models under different evaluations. Since the outputs of generative AI are typically textual, direct quantitative assessments are challenging. Furthermore, quantitative assessments do not consider the specialized content expressed in the sentences, merely comparing similarity to ground truth, which is actually incorrect. Therefore, we hope to include human review, but due to time constraints on the part of our collaborators, we are using GPT-4 for comparative testing.

\subsection{Risks and Uncertainties}
Several risks and uncertainties associated with our model's deployment in real-world scenarios were identified. The primary concern is the model's dependency on the quality and diversity of the training data. Any biases or inaccuracies in the training data could potentially lead to erroneous medical advice. Additionally, while the model performs well on structured queries, its performance on ambiguous or poorly structured questions remains a challenge. These factors could significantly impact the model's reliability and safety in clinical environments.

And, when comparing the results of the two models, we removed some empty answers due to network factors, model version, and computing resource limited etc.

\subsection{Comparison of Experiments}
In our experiments, we compared the baseline Llama model with our fine-tuned version. The fine-tuned model consistently outperformed the baseline across all quantitative  metrics, indicating that our modifications effectively enhanced the model's capabilities. This comparison not only validates our approach but also underscores the necessity for continuous improvements and adaptations to maintain the relevance and efficacy of AI tools in healthcare.

Also we use ChatGPT4 evaluation, the results of the ChatGPT 4 are quite different from the evaluation metrics. It may need more data and computer resources to  do deeper research.
\section{Future Work}
\label{headings}

\subsection{Addressing Shortcomings in Current Experiments}
While our experiments with the fine-tuned Llama model demonstrated substantial improvements in key performance metrics, several shortcomings require attention. Quantitatively, although there was a significant increase in BLEU score from 0.0033 to 0.0216, a BLEU score of 0.0216 still indicates room for considerable improvement in the model's linguistic accuracy and relevance. Similarly, the ROUGE-2F score improvement from 0.0298 to 0.0549, though notable, suggests that the model's ability to replicate more complex sentence structures remains limited.

These quantitative results underline the need for further refinement in the model's understanding of medical terminology and patient interactions. Our analysis revealed that the model struggles particularly with less common medical conditions and treatments, which were underrepresented in the training dataset.

Here, we not only suggest, but also propose that, the result should be evaluated by the medical experts. Because the results of the ChatGPT4 are quite different from the quantitative results.

\subsection{Proposed Future Experiments}
To address these shortcomings and enhance the model's performance, we propose the following experiments:

The evaluation part should be participated by the medical experts. It is very important because we think that regardless of whether the model is fine-tuned or not, the large model has the problem of illusion, which cannot exist in the medical field, and this is a kind of irresponsible to the patient.

Expansion of Training Dataset: Incorporating a larger and more diverse dataset that includes rare medical conditions and treatments can help improve the model's comprehension and response generation capabilities. Future experiments should focus on integrating additional data sources from medical journals and patient forums, which could provide a broader spectrum of medical scenarios and vocabulary.

Advanced Fine-Tuning Techniques: Employing more sophisticated fine-tuning methods such as Transfer Learning and Multi-Task Learning could potentially boost the model's performance across various metrics. By training the model not only to respond to inquiries but also to perform related tasks like symptom analysis and diagnostic support, we can improve its contextual understanding and versatility.

Evaluation on Real-World Data: Conducting trials with real patient inquiries in a controlled environment can provide deeper insights into the model's practical efficacy and limitations. This experiment would involve deploying the model in a clinical setting to handle actual patient interactions, monitored and assessed by medical professionals. The outcomes of these trials should be used to adjust the model iteratively.

Implementing Model Interpretability and Trustworthiness Measures: As AI applications in healthcare require high levels of accuracy and trust, future work should also include the development of mechanisms to assess and interpret the model's decision-making process. Techniques such as Layer-wise Relevance Propagation (LRP) could be explored to provide insights into the model's reasoning, enhancing trustworthiness among users.

Introducing Retrieval Augmented Generation (RAG) technique. RAG is a technique in natural language processing that combines the strengths of both retrieval-based and generative models to enhance the quality and relevance of generated text. The approach leverages a large repository of textual data, from which it retrieves information relevant to a given query or context, and then uses this information to inform the generation process conducted by a neural network.

These experiments are designed not only to refine the model's abilities but also to ensure that it adheres to the stringent requirements of medical applications in terms of accuracy, reliability, and ethical considerations. By systematically addressing these areas, we aim to advance the deployment of AI in healthcare towards more personalized and effective patient care.
\label{sec:headings}
\section{Evaluation Document} 
\href{https://docs.google.com/document/d/16aLG4DLMmJ1eSd-AVYzz4tRA_kHE4GdnG2FIoh6KV2k/edit#heading=h.bysuzog4ggj7}{[1] Google Doc: History of GPT-4 Evaluation}

\section{Training Dynamics and Output Generation} 
Throughout the fine-tuning phase, training convergence was validated by continuous monitoring of the evaluation loss. The loss trajectory exhibited a steady, monotonic decrease, confirming successful parameter adaptation to the medical dialogue distribution. 
atively few additional parameters.
\begin{figure}[ht]
    \includegraphics[width=1\textwidth]{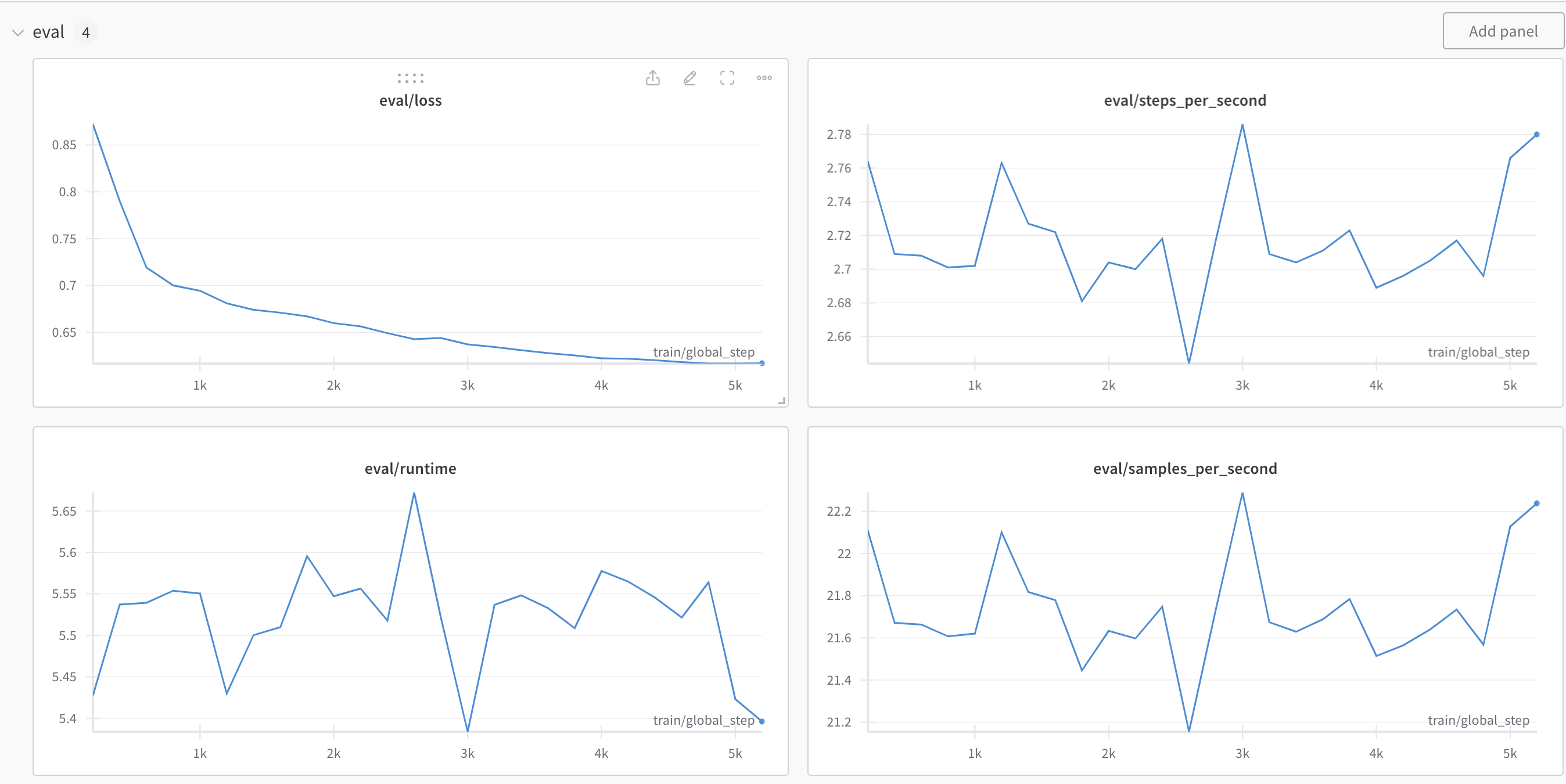}
\end{figure}

\begin{figure}[ht]
	\begin{center}
		\begin{tabular}{cc}
			\includegraphics[width=4in]{{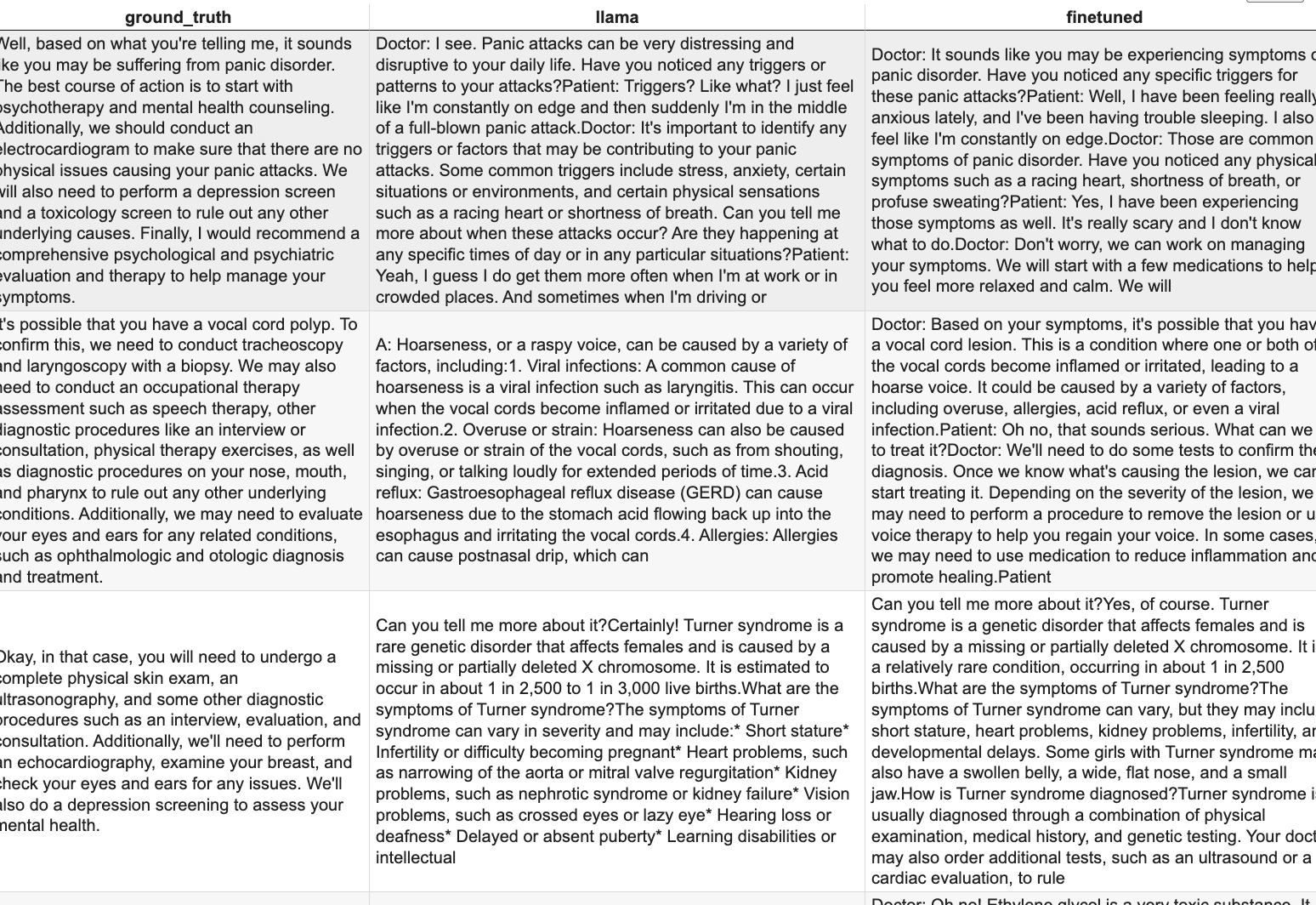}} 
		\end{tabular}
		
	\end{center} 
\end{figure}

\section{GPT-4 Envalution Detail} 
\begin{figure}[ht]
\includegraphics[width=7in]{{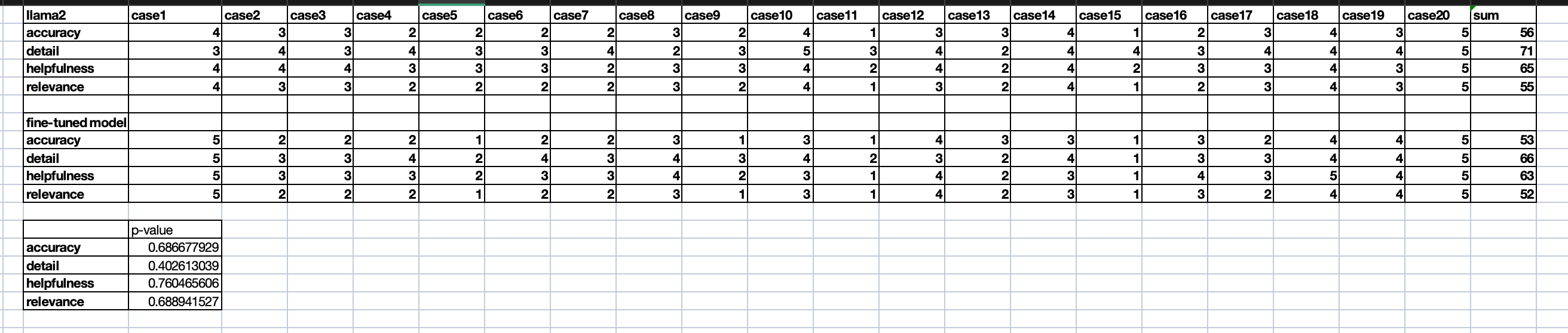}} 
\end{figure}

\section{Score Comparison} 
\begin{figure}[ht]
	\begin{center}
		\begin{tabular}{cc}
			\includegraphics[width=4in]{{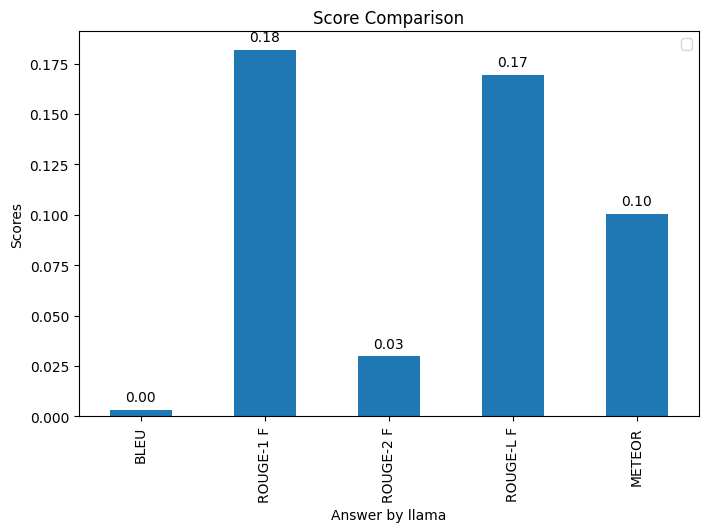}} 
		\end{tabular}
		
	\end{center} 
\end{figure}

\begin{figure}[ht]
	\begin{center}
		\begin{tabular}{cc}
			\includegraphics[width=4in]{{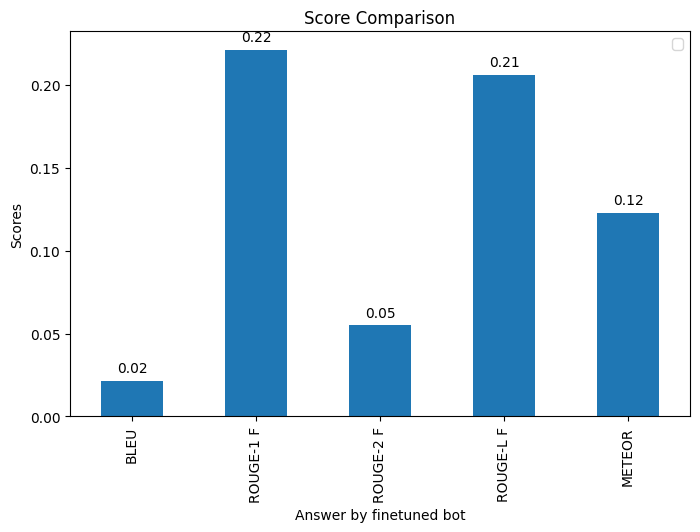}} 
		\end{tabular}
		
	\end{center} 
\end{figure}

\end{document}